\documentclass[letterpaper, 10 pt, conference]{ieeeconf}  % Comment this line out if you need a4paper

\IEEEoverridecommandlockouts                              % This command is only needed if 
\usepackage{graphicx} % for pdf, bitmapped graphics files
\usepackage{amsmath} % assumes amsmath package installed
\usepackage{amssymb}  % assumes amsmath package installed
\usepackage[hyphens]{url}
\usepackage{hyperref}
\usepackage{setspace}
\usepackage{caption}
\usepackage{subcaption}
\usepackage{algorithm2e}
\usepackage{bm}
\usepackage{comment}

\RestyleAlgo{ruled}
\SetKwComment{Comment}{/* }{ */}

\title{\LARGE \bf
Towards Probabilistic Causal Discovery, Inference \& Explanations \\
for Autonomous Drones in Mine Surveying Tasks
}

\author{Ricardo Cannizzaro$^{1*}$, Rhys Howard$^{1*}$, Paulina Lewinska$^{2}$, and Lars Kunze$^{1}$% <-this % stops a space
\thanks{
This work was supported by the ASUMI project, the Assuring Autonomy International Programme, EPSRC project RAILS (grant reference: EP/W011344/1) and the Australian Defence Science and Technology Group.}
\thanks{$^{1}$\,Ricardo Cannizzaro, Rhys Howard and Lars Kunze are with the Oxford Robotics Institute,
Department of Engineering Science, University of Oxford, 17 Parks Road, Oxford OX1
3PJ, UK. Please use rhyshoward@live.com for correspondence.}%
\thanks{$^{2}$\,Paulina Lewinska is with the Department of Computer Science, University of York, Heslington, York YO10 5DD, UK.}%
\thanks{* Cannizzaro and Howard are identified as joint lead authors of this work. }% <-this % stops a space
}

\begin{document}

\maketitle
\thispagestyle{empty}
\pagestyle{empty}

%%%%%%%%%%%%%%%%%%%%%%%%%%%%%%%%%%%%%%%%%%%%%%%%%%%%%%%%%%%%%%%%%%%%%%%%%%%%%%%%

\begin{abstract}
% Set the context first - why is this important?
%There is growing interest to use micro uncrewed aerial vehicles (UAVs), i.e. drones, for mine frontier surveying, to supplement human labour in underground mining applications. 
% What is the problem here?
%However, these environments are challenging for autonomous robots because of their uncertain, dynamic, partially-observable, partially-controllable, cluttered, and causally-complex characteristics \textemdash\ critical challenges that must be solved for these autonomous systems to be used in a safe, reliable, and trustworthy manner.

% What are we proposing to do about it? And how?
Causal modelling offers great potential to provide autonomous agents the ability to understand the data-generation process that governs their interactions with the world. Such models capture formal knowledge as well as probabilistic representations of noise and uncertainty typically encountered by autonomous robots in real-world environments. Thus, causality can aid autonomous agents in making decisions and explaining outcomes, but deploying causality in such a manner introduces new challenges. 
%Probabilistic causal modelling and inference holds great potential to provide autonomous agents with the ability to understand the data-generation process that governs their interactions with the world, capturing both formal knowledge representation and a probabilistic representation of the sources of noise and uncertainty typically encountered by autonomous robots in real-world environments. These causal models can be constructed in a hybrid manner by combining domain knowledge with offline and online learning.
Here we identify challenges relating to causality in the context of a drone system operating in a salt mine. Such environments are challenging for autonomous agents because of the presence of confounders, non-stationarity, and a difficulty in building complete causal models ahead of time.
To address these issues, we propose a probabilistic causal framework consisting of: causally-informed POMDP planning, online SCM adaptation, and post-hoc counterfactual explanations. Further, we outline planned experimentation to evaluate the framework integrated with a drone system in simulated mine environments and on a real-world mine dataset.

\end{abstract}

%%%%%%%%%%%%%%%%%%%%%%%%%%%%%%%%%%%%%%%%%%%%%%%%%%%%%%%%%%%%%%%%%%%%%%%%%%%%%%%%

\section{INTRODUCTION}
%With recent advances in micro uninhabited aerial vehicles (UAVs), i.e. drones, autonomous UAVs are being increasingly used to augment human labour in a variety of industrial tasks.
%(Bridging sentence to go here). 
When building autonomous systems it is critical to incorporate an understanding of how system and environment variables interact when deciding how to act. 
Structural causal models (SCMs) can represent such interactions; however, these are typically either defined by experts or learned via offline causal discovery, and cannot robustly adapt to unanticipated environmental conditions. 
We explore the challenges posed by this in the context of real-world drone usage in mine survey tasks. 
This work has three main contributions. First, we describe the drone mine survey task and key challenges. Second, we propose a probabilistic causal framework consisting of: causally-informed confounder-robust POMDP planning; online SCM adaptation to unanticipated environmental phenomen; and post-hoc counterfactual explanations. Finally, we discuss planned experiments to evaluate our proposed framework on drone missions in simulated mine environments and on a real-world mine dataset.

\section{PROBLEM DESCRIPTION} \label{sec:problem_definition}

\begin{figure}
    \centering
    \begin{subfigure}[b]{0.55\linewidth}
    \centering
    \includegraphics[width=\linewidth]{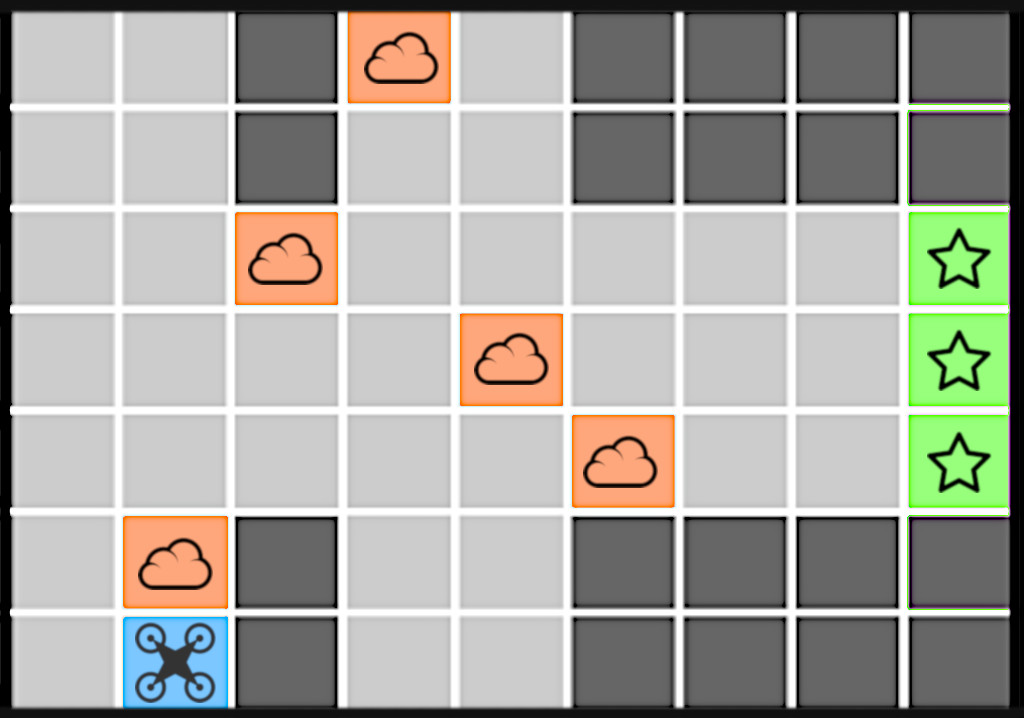}
    \caption{}
    \label{subfig:gridworld}
    \end{subfigure}
    \hfill
    \begin{subfigure}[b]{0.4\linewidth}
    \centering
    \includegraphics[width=\linewidth]{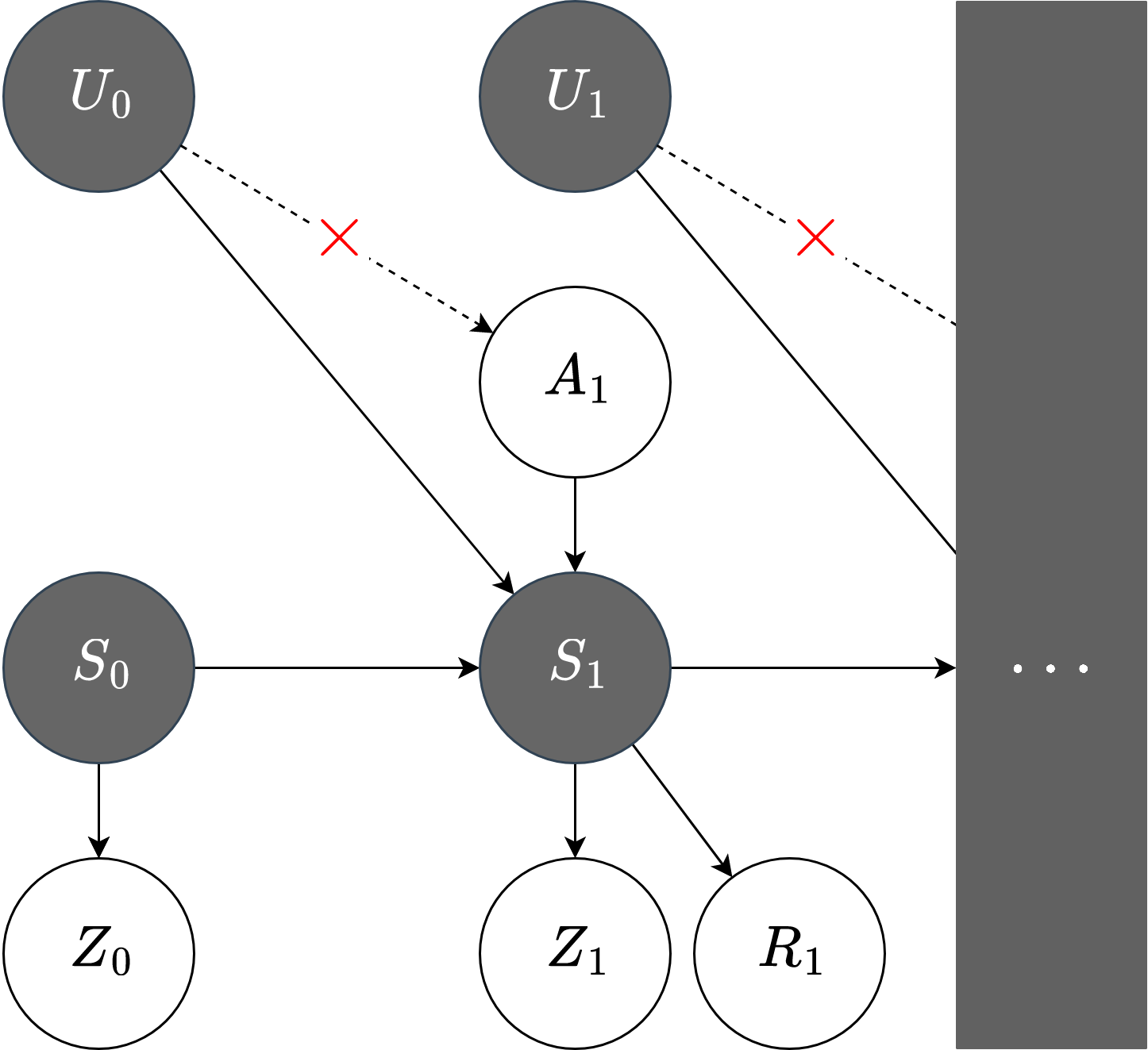}
    \caption{}
    \label{subfig:scm_ucpomdp}
    \end{subfigure}
    \caption{(a) Depiction of a mine environment in grid-world format. (b) Depiction of a SCM-POMDP with unobserved confounded (SCM-UCPOMDP) with interventions on the drone's actions, illustrated by the red crosses.}
\end{figure}

%As previously established, the scenario we are aiming to tackle here concerns the autonomous operation of a drone in a mining environment. Here we detail the exact specifics of the environment, drone task and provide a formal definition of the challenges in terms of causality theory.

\subsection{Drone Mine Survey Application}
% introduce the survey task
To aid in evaluating the efficiency of the mining process, volumetric scans can be taken pre- and post-excavation. It is advantageous to carry this process out via drone to avoid human entry into hazardous areas. Typically these drones are manually operated remotely by a pilot, however the subterranean conditions greatly increase the risk of drone signal loss. As such, it is desirable to have an autonomous framework that upon pilot signal loss navigates the drone out of the hazardous excavated area and lands it safely. 

\subsection{Environmental Challenges}
Two potential sources of deviation from the established drone-task causal model may be present: wind from ventilation systems; and salt dust either suspended in the air or sticking to the drone, which affects motors and sensors. Further, the environment is only partially observable and may change over time due to shifting salt material.

\subsection{Problem Definition}

%In this work we address three key problems to enable the safe and trustworthy use of autonomous drones in mining applications.
%
\subsubsection{Confounder-Robust Planning}
First, we address the problem of a drone autonomously navigating to a designated return-to-home region, determining a safe landing location, and landing following remote control signal loss.
%and disarming for collection by a human, in response to an identified pilot remote control signal loss failsafe event. 
Specifically, the autonomous drone must plan and execute safe flight trajectories that satisfy collision, kinematic, and mission safety constraints under uncertain, complex, partially-observable, and dynamic mine conditions. A depiction of this task as a grid-world is shown in Fig. \ref{subfig:gridworld}.
To achieve the aforementioned task, we propose the use of the CAR-DESPOT confounder-robust online POMDP planner and an assumed fixed model built from expert domain knowledge, as per our earlier work \cite{cannizzaro2023cardespot}. In Fig. \ref{subfig:scm_ucpomdp} we illustrate the interventions on SCM-UCPOMDPs utilised by CAR-DESPOT.
\subsubsection{Online Causal Model Adaptation}
Second, we tackle adapting the drone's causal model online in response to observations of the environment that deviate from said causal model. In the context of a salt mine, the wind conditions may change quickly and at random due to activity at air dams, while the exact interactions of the salt dust with motors and sensors is uncertain and difficult to model in advance. Thus, in order for the causal model to remain informative it must be adapted online while the drone operates.
\subsubsection{Post-Hoc Counterfactual Explanations}
Finally, we address the problem of creating a capability for the autonomous system to generate explanations for sensor measurements, decisions, and observed mission outcomes, in response to queries issued by a human via a human-robot interaction system following the robot deployment. 

% Not enough space, so removed for the time being
% \subsection{Formal Causal Definition}
% TODO

\section{RELATED WORK}

\begin{figure}
    \centering
    \includegraphics[width=0.9\linewidth]{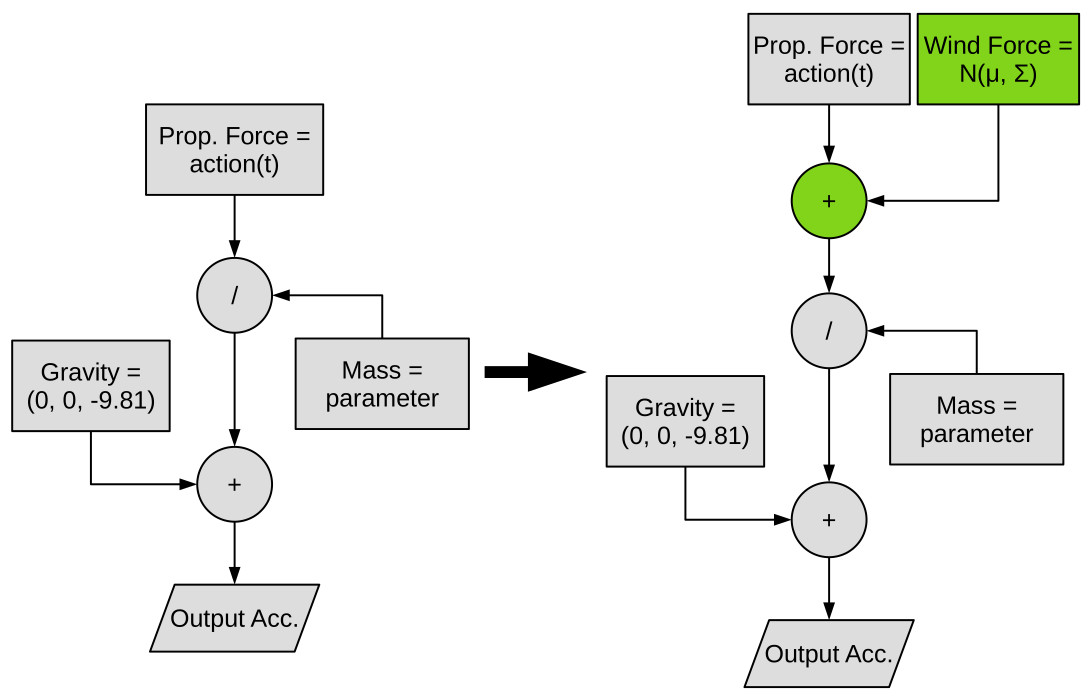}
    \caption{Illustration of part of an SCM before and after online adaptation. The adaptation here involves adding a conditional exogenous variable to represent the force of a newly present gust of wind which is added to the existing force from propellors via an endogenous sum variable. The new wind force and sum variables are both indicated in green.}
    \label{fig:scm_adaptation}
\end{figure}

The application of causality to robotics has received an increasing level of attention in recent years \cite{hellstrom2021relevance}; previous work has used it to inform planning and control and offer an increased level of transparency regarding the underlying data generation process. 

Probabilistic robot planning under uncertainty is a well-established field \cite{kurniawati2022pomdpreview}; however, only recently have causal formulations been adopted for robot planning problems \cite{diehl2022causalexplainpredictprevent}. More recently it has been demonstrated that utilising an SCM-based POMDP can result in more robust offline planning in environments with confounders \cite{cannizzaro2023cardespot}.
%More recently in \cite{cannizzaro2023cardespot}, authors introduce a causal SCM-based POMDP model formulation and propose an extended online POMDP planner that is demonstrated to produce higher performing policies in a toy problem of robot planning in a confounded environment. However, this method has yet to be demonstrated to solve more realistic robot planning problems.

A recent technique to learn causal models online is causal Reinforcement Learning (RL) \cite{bareinboim2015bandits,gasse2021causal}. RL-based approaches balance learning the causal model via interactions with the environment against exploiting the causal model to accomplish the agent's objective. A similar technique is continual causal learning. For example, the approach of Castri \textit{et al.} \cite{castri2023continual} only has the agent intervene upon a discrepancy being found between subsequent observation-based causal models.
%
% Commenting out since we're gonna cite this in the proposed architecture section
%Some approaches deliberately eschew online discovery and focus upon learning and utilising causal model parameterisations for a hand build SCM \cite{cannizzaro2023cardespot}. While this does restrict such methods to purely training off observations, this has the benefit that it allows such methods to be used in risky applications, where experimentation is not feasible.
%

There is a rich body of literature in the area of observation-based causal discovery \cite{assaad2022survey}.
% but not necessarily how they ought to be utilised. 
However, results from recent works indicate that existing causal discovery techniques score low in precision when working with agent-related data \cite{howard2023evaluating} due to non-stationarity and limited data. Thus, there is a growing interest in addressing these challenges \cite{huang2019causal}.

In the trustworthy autonomous systems literature, recent works aim to explain the perceptions, actions, and outcomes of robots post-deployment \cite{kunze2022embodiedquestionanswering}. However, few use a causal formulation of explainability and to our knowledge no prior work has employed counterfactual-based methods to generate explanations. For example, Diehl \& Ramirez-Amaro \cite{diehl2022causalexplainpredictprevent} learn a causal Bayesian network from simulation data to generate \emph{contrastive explanations}, but this may prove unreliable if the outcome is sufficiently impacted by random variables \cite{gerstenberg2022causaljudgements}.
%For example, in \cite{diehl2022causalexplainpredictprevent}, authors learn a causal Bayesian network from simulation data to generate contrastive explanations \textemdash\ a method in which outcomes are explained only by changes in initial conditions. However, contrastive explanations may diverge from counterfactual explanations when the outcome is impacted by another random variable, and, further, a recent study showed counterfactual explanations to have better alignment with human causal judgements \cite{gerstenberg2022causaljudgements}.

\section{PROPOSED ARCHITECTURE}

\begin{algorithm}[t]
\caption{Online SCM Adaption Loop}\label{alg:scm_adaptation}
\small
\KwData{$\mathcal{M} = \langle U, V, F, P(U) \rangle,\ P(\mathcal{M}),\ N \in \mathbb{N}$}
\KwResult{$\hat{\mathcal{M}}$}
$\hat{\mathcal{M}} \gets \mathcal{M}$\;
$\boldsymbol{M} \gets \{ \mathcal{M} \}$\;
\For{$t\gets0$ \KwTo $\infty$}{
\Comment{Gather observational data}
$\mathcal{D}_t \gets GetData()$\;
\Comment{Generate a set of altered SCMs}
$\boldsymbol{M} \gets \boldsymbol{M} \cup GenAltSCM(\boldsymbol{M})$\;
$P(\mathcal{D}_{0..t}) \gets 0$\;
\For{$\mathcal{M}^\prime \in \boldsymbol{M}$}{
    \Comment{Approximate the likelihood of the data for a given SCM}
    $P(\mathcal{D}_t\, |\,\mathcal{M}^\prime) \gets Likelihood(\mathcal{D}_t, \mathcal{M}^\prime, \mathcal{D}_{t-1})$\;
    $P(\mathcal{D}_{0..t},\,\mathcal{M}^\prime) \gets P(\mathcal{D}_t\, |\,\mathcal{M}^\prime) P(\mathcal{M}^\prime\, |\,\mathcal{D}_{0..t-1})$\;
    $P(\mathcal{D}_{0..t}) \gets P(\mathcal{D}_{0..t}) + P(\mathcal{D}_{0..t},\,\mathcal{M}^\prime)$\;
}
\For{$\mathcal{M}^\prime \in \boldsymbol{M}$}{
    $P(\mathcal{M}^\prime\, |\,\mathcal{D}_{0..t}) \gets P(\mathcal{D}_{0..t},\,\mathcal{M}^\prime)\ /\ P(\mathcal{D}_{0..t})$\;
}
\Comment{Keep the N most probable SCMs}
$\boldsymbol{M} \gets \boldsymbol{M} \cap MaxProbs(\boldsymbol{M}, N)$\;
\Comment{Use the most probable SCM as the current best estimate}
$\hat{\mathcal{M}} \gets MaxProb(\boldsymbol{M})$\;
}
\end{algorithm}

\subsection{Causally-Informed POMDP Planning}
In addition to possessing an SCM model, a causal inference and planning framework is required for autonomous drone sensing and decision-making.
Our proposed framework will utilise the causally-informed CAR-DESPOT POMDP planner formulated by Cannizzaro \& Kunze \cite{cannizzaro2023cardespot}, which will be used to represent the action-selection process of the drone during a given time-step as an intervention in the SCM causal world model.
Through the SCM informing the planner of drone-environment dynamics and by intervening on action selection (e.g., via an override mechanism) the system becomes more aware of the influence of random variables, and more robust to environmental confounders.
%We anticipate this will permit the drone to generate better performing policies because: 1) the causal relationships that govern the drone-environment dynamics are considered during planning; and, 2) the ability to intervene on action selection permits the drone agent to plan in a deliberative manner and thus avoid influence from confounding elements in the environment that could otherwise cause other drone sub-system components to induce reactive drone behaviours (e.g., collision avoidance) that may lead to catastrophic failures.

\subsection{Online SCM Adaptation}

% Points to hit
% We want to phrase Online SCM discovery as an optimization problem, similar to SLAM
% We want to explore existing optimisation methods, such as genetic algorithms and particle filter based methods, but apply it to incremental SCM structure learning
%The difficulty of online causal discovery in a hazardous environment such as a mine, is that experimentation is risky, thus limiting the usage of reinforcement learning based techniques that typically fill this niche. Observation-based techniques on the other hand are almost entirely offline, requiring large amounts of pre-collected data in order to make up for the lack of ability to intervene while learning.

%We propose that by drawing parallels between this challenge and Simultaneous Localisation And Mapping (SLAM), we can develop a system that can adapt to new environmental conditions as they are encountered. 
We propose that adapting an SCM online should be tackled as an optimisation problem, following in the footsteps of online methods in other fields (e.g., SLAM). 
%Our work will consider techniques such as particle filters and more generally genetic algorithms, which bear resemblance to the methods utilised by existing score-based causal discovery approaches \cite{assaad2022survey}.
Huang \textit{et al.} \cite{huang2019causal} provide an example of particle filter use, to find linear relationships between a set of time-indexed variables under non-stationary conditions. However, their work relies upon possessing sufficient data, and expects the non-stationarity to follow a prescribed pattern, making it unsuitable for fast adaptation to new phenomena. 

We aim to simplify the problem by not considering time-lagged relations, fixing the nature of noise to Gaussian noise, and beginning with an initial predefined SCM. Through these simplifications we aim to make SCM adaptation feasible in a matter of seconds such that it can operate in real-time. We depict the type of adaptations we hope to achieve in Fig. \ref{fig:scm_adaptation}.
%Returning to our analogy with SLAM, we aim to carry out the creation and destruction of variables and edges as a separate step to the parameterisation of exogenous variables.
%Here we liken the creation of the SCM to the online creation of a map during SLAM, while the aforementioned parameterisation of exogenous variables within the SCM is likened to localisation within said map.

Meanwhile we provide a high-level algorithm of the planned approach in Algorithm \ref{alg:scm_adaptation}. As previously mentioned, the approach bears resemblance to a particle filter, creating alternations to the SCM and evaluating the likelihood of observed data given each SCM before updating the belief over SCMs by utilising the Markovian nature of the drone-environment states. The aforementioned likelihood will be approximated via Monte Carlo sampling methods to generate distributions for observed variables, e.g. \cite{kloek1978bayesian}.

\subsection{Post-Hoc Counterfactual Explanations}
Counterfactual inference permits an agent to simulate what would have happened, had certain things in past been different to the observed reality. This is done by intervening on target variables while maintaining the inferred distributions of other variables.
Our proposed approach to generating post-hoc counterfactual explanations of a drone deployment will use twin-world counterfactual algorithm based methods 
%to compute the probability of necessity and sufficiency 
to search for the random variable with the highest probability of being a necessary and sufficient cause for the observed outcome. 
Explanation variables will include key variables governing the sensing, decision-making, and actuating variables relevant to the drone; and environmental factors.
%process of the drone, including environmental conditions, perceptions, decisions made by the planner, and stochastic outcomes.
%In addition to interrogating system performance post-deployment, the identified causes of drone system failures can also be used to further improve the SCM causal model and/or the decision-making process to incrementally improve the future system performance.
% By comparing the distribution of outcomes inferred under these interventions with the observed outcome we can approximate a potential cause for failure conditions that can aid in taking corrective action and refining future iterations of the system.

\section{EVALUATION SETUP}
% Need to put a few lines saying what evaluation we want to do with the sim environments that will evaluate the method's performance on the problem, qualitatively and/or quantatively

\begin{comment}
\begin{figure}
    \centering
    % top row
    \includegraphics[width=0.32\linewidth]{figures/inside_blender.jpg}
    \includegraphics[width=0.32\linewidth]{figures/inside_real_balanced.jpg}
    \includegraphics[width=0.32\linewidth]{figures/real_mine_mesh_crop2.png}\\[1mm]
    % bottom row
    \includegraphics[width=0.32\linewidth]{figures/overhead_blender.jpg}
    \includegraphics[width=0.32\linewidth]{figures/outside_gazebo.jpg}
    \includegraphics[width=0.32\linewidth]{figures/real_overhead_blender.png}
    \caption{Drone mine survey environments. Top row L-R: interior of simulation mock mine, real-world mock mine, and real mine. Bottom row L-R: exterior and overhead of simulation mock mine, overhead of real mine.}
    \label{fig:environments}
\end{figure}
\end{comment}

\begin{figure}
    \centering
    % top row
    \includegraphics[width=0.46\linewidth]{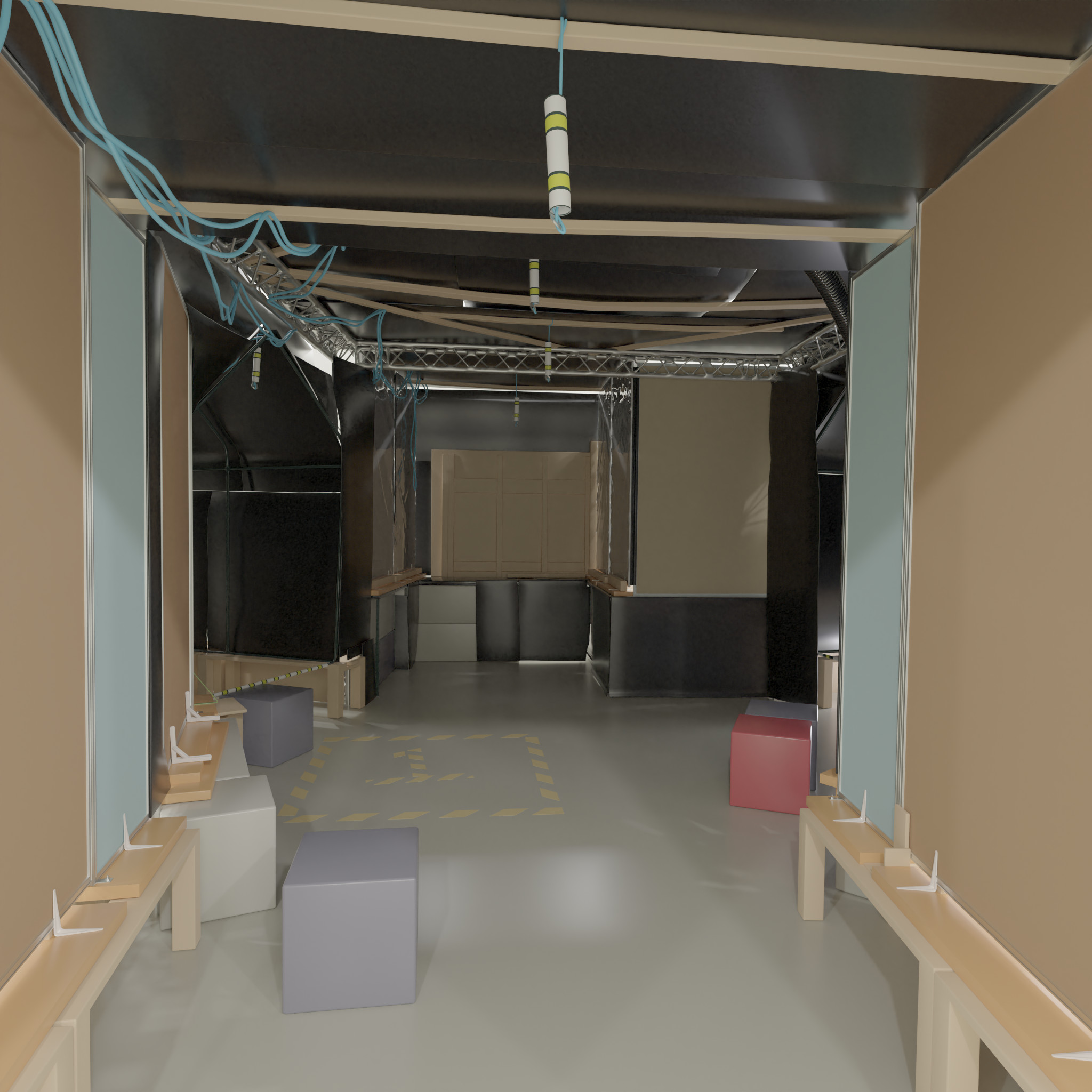}
    \includegraphics[width=0.46\linewidth]{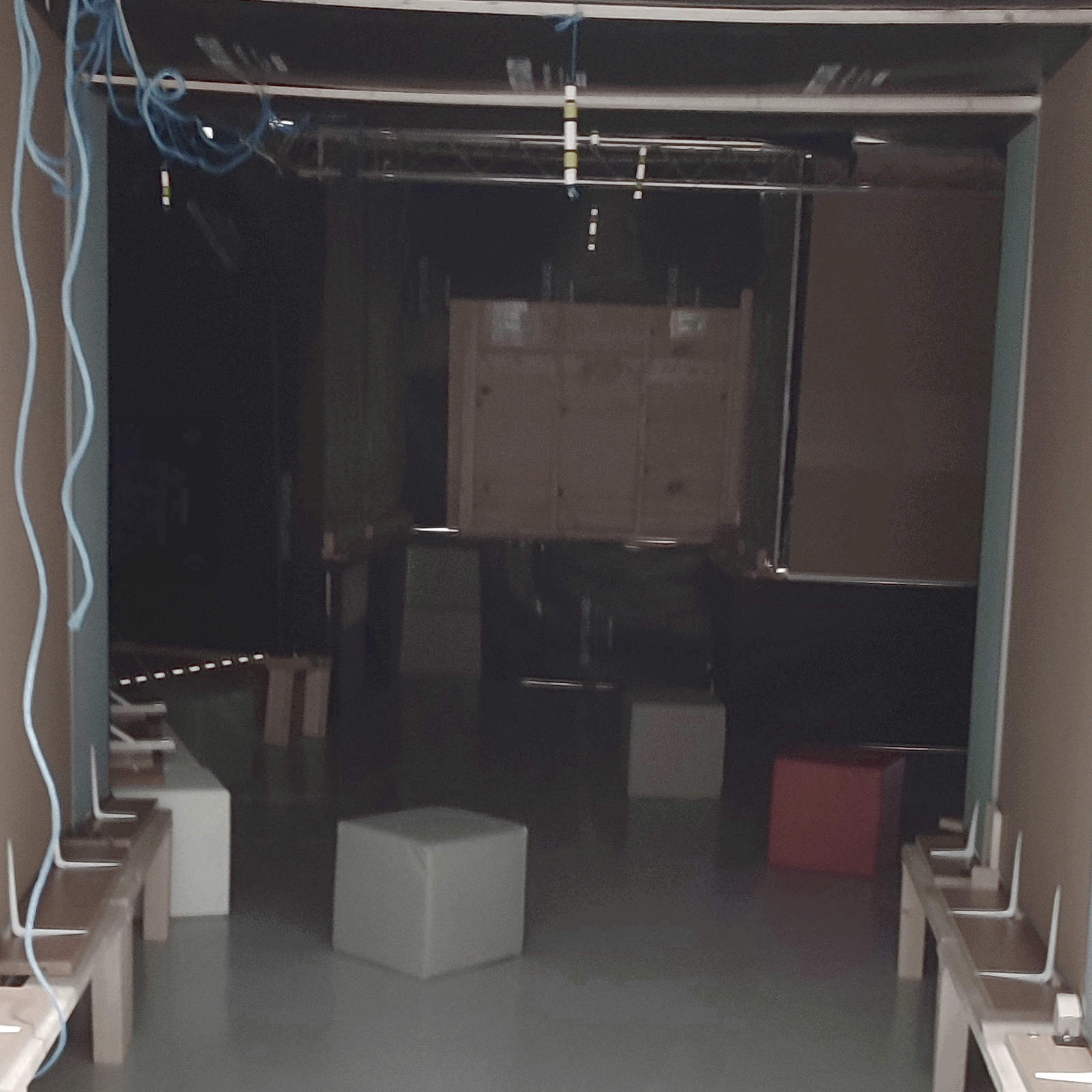}\\[1mm]
    % middle row
    \includegraphics[width=0.46\linewidth]{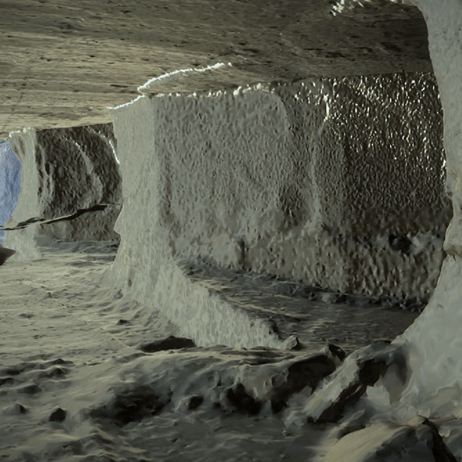}
    \includegraphics[width=0.46\linewidth]{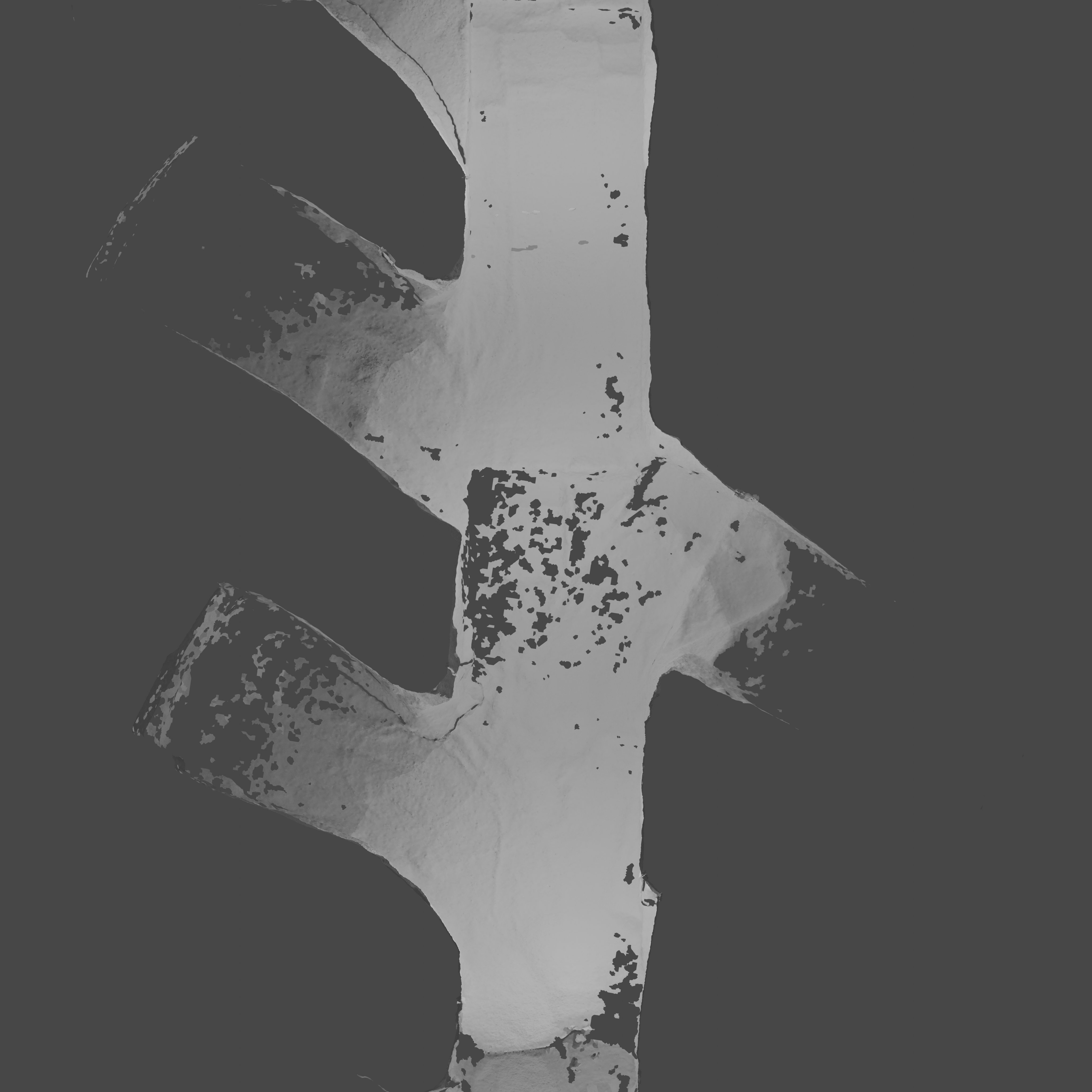}\\[1mm]
    % bottom row
    \includegraphics[width=0.46\linewidth]{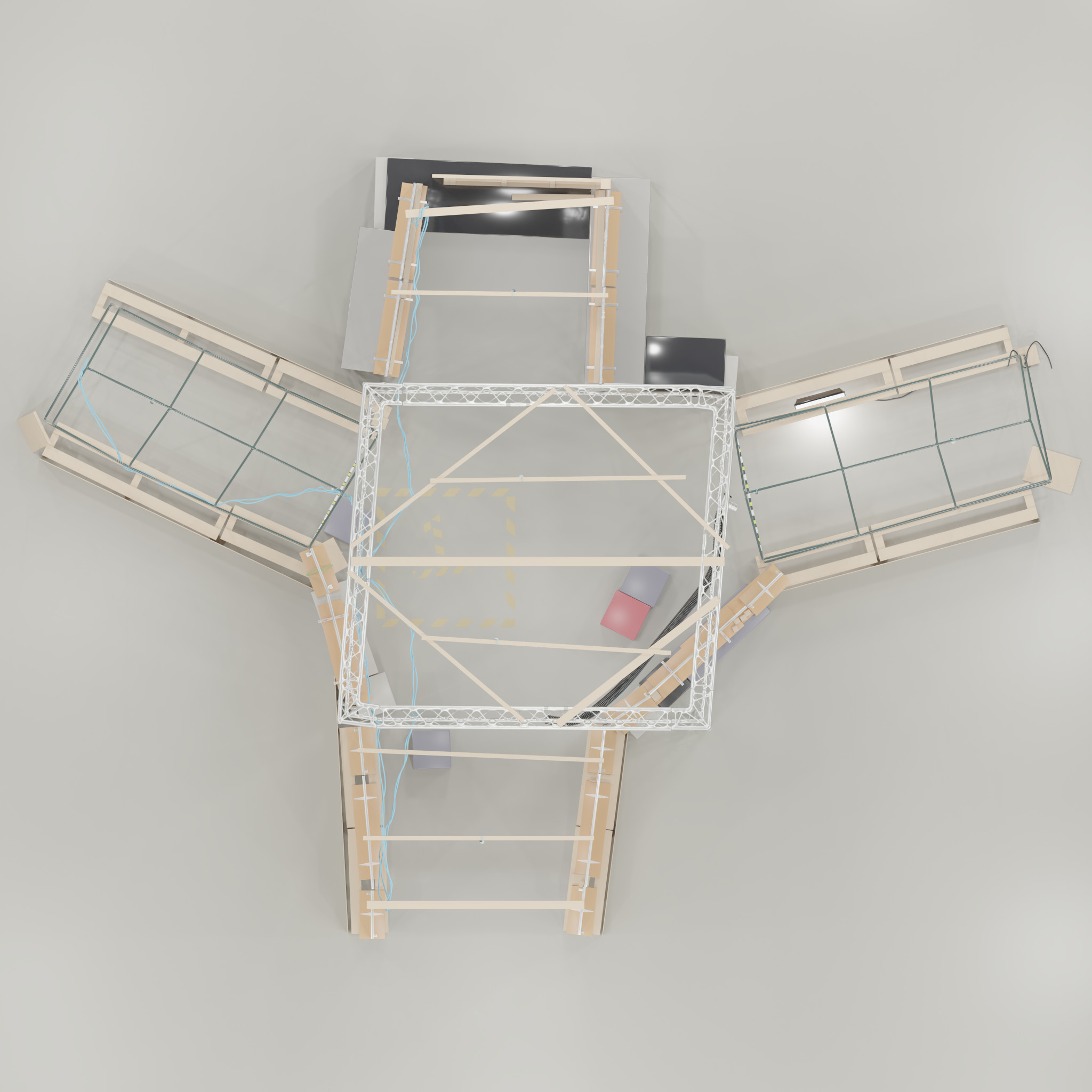}
    \includegraphics[width=0.46\linewidth]{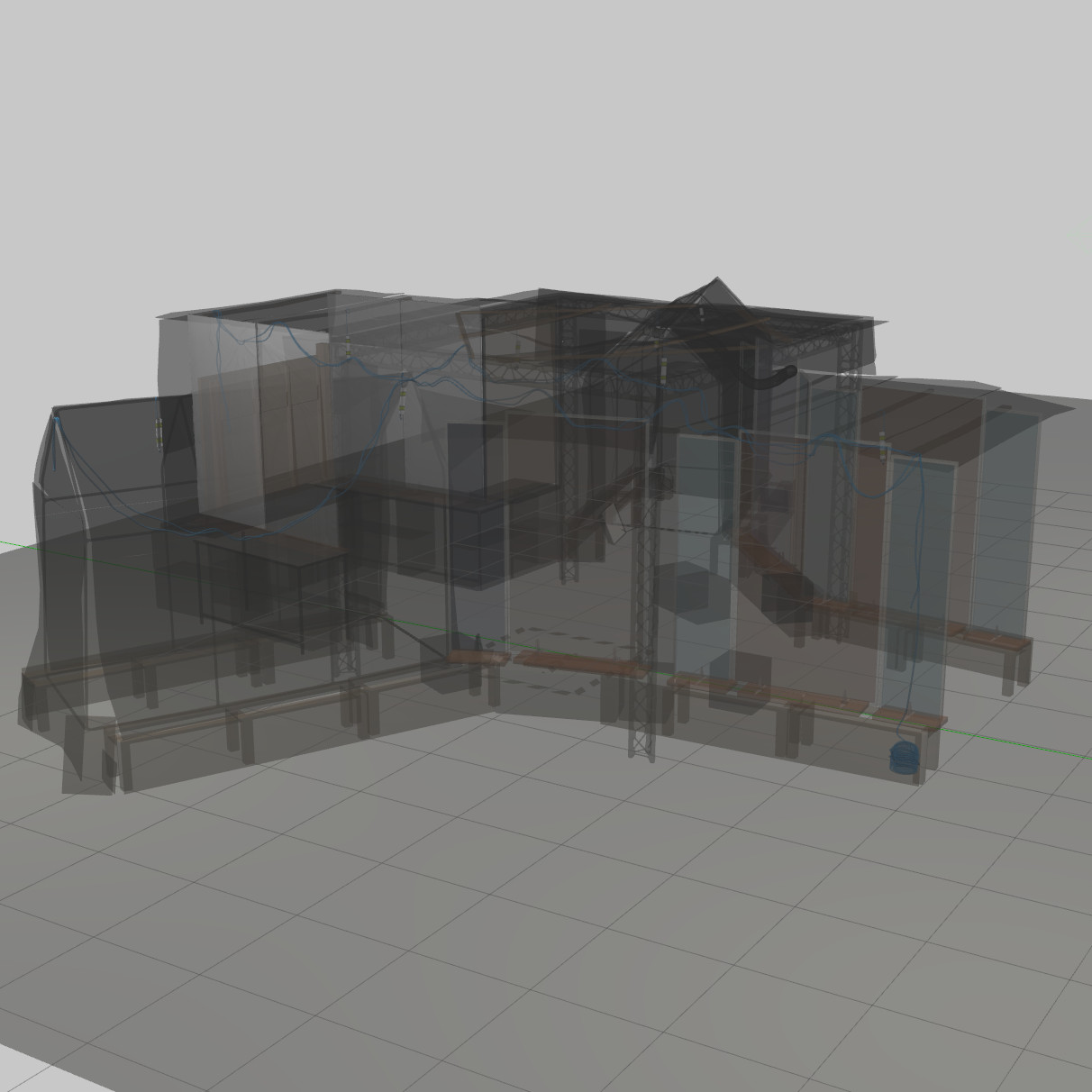}
    \caption{Drone mine survey environments. Top row L-R: interior of simulation mock mine and real-world mock mine. Middle row L-R: interior and overhead of real mine. Bottom row L-R: overhead and exterior of simulation mock mine.}
    \label{fig:environments}
\end{figure}

% Commented to save space
\subsection{Environments}
Here we describe one simulated test environment and two real-world test environments, as depicted in Fig. \ref{fig:environments}.

\subsubsection{Simulation Mock Mine}
A simulation-based mock mine, for rapid testing. Permits the emulation of environmental factors that are infeasible in the real-world version. This environment is made available via a Git repository\footnote{Simulation Mock Mine Environment: \url{https://github.com/cognitive-robots/aaip-mine-envs}}.

\subsubsection{Real-World Mock Mine}
A constructed mock mine setup in an indoor flight lab, for evaluation on a real drone. Contains a subset of the salt-mine environmental factors.

\subsubsection{Real Salt Mine Dataset}
In collaboration with an industry partner, we will evaluate the proposed architecture using a dataset generated in a real-world active salt mine, to assess against true mine conditions.

\subsection{Hardware \& Software}
The target hardware system for the mock mine environments is a PX4 Vision V1 drone equipped with a Structure Core depth camera and Up Core companion computer 
\footnote{PX4 Vision Kit: \url{https://docs.px4.io/main/en/complete_vehicles/px4_vision_kit.html}}.
The companion computer will interface with the flight control unit via MAVLink and the causal framework via ROS. A simulated 3D LIDAR will generate synthetic LIDAR measurements in the mock mine environments. Conversely, the existing real mine dataset was generated using an Elios 3 drone equipped with a camera, 3D LIDAR, and NVIDIA Xavier companion computer.

\subsection{Experiments}
The initial evaluation metric of the planning and SCM adaptation components of the framework will be the number of successful return-to-home missions, with randomised initial drone positions. Further metrics will also be considered.

Counterfactual explanations will be evaluated by: 1) using flight data recorded in the experimentation above to qualitatively compare system explanations to ground truth data; and, 2) using data generated via additional missions in simulation, in which a failure mode is intentionally induced, to qualitatively compare the accuracy of the system-identified highest-probability cause variable with the true failure cause.
\begin{spacing}{0.9}
\fontsize{8pt}{9pt}\selectfont
\bibliographystyle{IEEEtran.bst}
\bibliography{references.bib}
\end{spacing}

\end{document}